\documentclass{article}
\usepackage{geometry}
\PassOptionsToPackage{numbers, sort&compress}{natbib}
\usepackage{natbib}

\usepackage[utf8]{inputenc} 
\usepackage[T1]{fontenc}    
\usepackage{hyperref}       
\usepackage{url}            
\usepackage{booktabs}       
\usepackage{amsfonts}       
\usepackage{nicefrac}       
\usepackage{microtype}      
\usepackage{subcaption}
\usepackage{algorithm}
\usepackage{algorithmicx}
\usepackage{algpseudocode}
\usepackage{authblk}

\usepackage{xifthen}

\newcommand{\ifequals}[3]{\ifthenelse{\equal{#1}{#2}}{#3}{}}
\newcommand{\case}[2]{#1 #2} 
\newenvironment{switch}[1]{\renewcommand{\case}{\ifequals{#1}}}{}

\newcommand{\secwithdepth}[2]{
    \begin{switch}{#1}
        \case{0}{\chapter{#2}}
        \case{1}{\section{#2}}
        \case{2}{\subsection*{#2}}
        \case{3}{\subsubsection*{#2}}
        \case{4}{\paragraph{#2}}
    \end{switch}
}

\newcommand{\secwithcounter}[3]{
    \secwithdepth{\the\numexpr\value{#1}+#2\relax}{#3}
}

\usepackage{tikz}
\usetikzlibrary{shapes,decorations}
\usetikzlibrary{positioning}

\usepackage{amssymb}
\usepackage{nicefrac}       
\usepackage{amsfonts}       
\usepackage{amsmath}
\usepackage{amsthm}

\usepackage{MnSymbol}
\usepackage{mathtools}

\usepackage{siunitx}
\newcommand{\norm}[1]{\lVert#1\rVert}
\newcommand{\abs}[1]{\left\lvert#1\right\rvert}

\DeclareMathOperator*{\argmin}{\arg\!\min}

\DeclareMathOperator*{\E}{\mathbb{E}}

\DeclareMathOperator{\R}{\mathbb{R}}

\usepackage{mathtools}

\theoremstyle{definition}

\DeclareMathOperator{\DAG}{\mathbb{D}}

\makeatletter
\def\set@curr@file#1{\def\@curr@file{#1}} 
\makeatother

\newcommand{\method}{NOTMAD}
\newcommand{\methodfull}{\underline{NOT}EARS-optimized \underline{M}ixtures of \underline{A}rchetypal \underline{D}AGs}
\newcommand{\methodparens}{(\method)}
\renewcommand{\P}{\mathbb{P}}

\begin{document}

\title{\method: Estimating Bayesian Networks with Sample-Specific Structures and Parameters}

\author[1, 2]{Ben Lengerich\thanks{\texttt{blengeri@mit.edu}}}
\author[3]{Caleb Ellington}
\author[4]{Bryon Aragam}
\author[3, 5]{Eric P. Xing}
\author[1, 2]{Manolis Kellis}

\affil[1]{Massachusetts Institute of Technology}
\affil[2]{Broad Institute of MIT and Harvard}
\affil[3]{Carnegie Mellon University}
\affil[4]{University of Chicago}
\affil[5]{Mohamed bin Zayed University of Artificial Intelligence}

\maketitle

\begin{abstract}
Context-specific Bayesian networks (i.e. directed acyclic graphs, DAGs) identify context-dependent relationships between variables, but the non-convexity induced by the acyclicity requirement makes it difficult to share information between context-specific estimators (e.g. with graph generator functions). For this reason, existing methods for inferring context-specific Bayesian networks have favored breaking datasets into subsamples, limiting statistical power and resolution, and preventing the use of multi-dimensional and latent contexts. 
To overcome this challenge, we propose \methodfull~(\method). 
\method~models context-specific Bayesian networks as the output of a function which learns to mix archetypal networks according to sample context. 
The archetypal networks are estimated jointly with the context-specific networks and do not require any prior knowledge. 
We encode the acyclicity constraint as a smooth regularization loss which is back-propagated to the mixing function; in this way, \method~shares information between context-specific acyclic graphs, enabling the estimation of Bayesian network structures and parameters at even single-sample resolution. 
We demonstrate the utility of \method~and sample-specific network inference through analysis and experiments, including patient-specific gene expression networks which correspond to morphological variation in cancer.
\end{abstract}

\section{Introduction}
\label{sec:introduction}

\begin{figure*}[t]
    \centering
    \includegraphics[width=\textwidth]{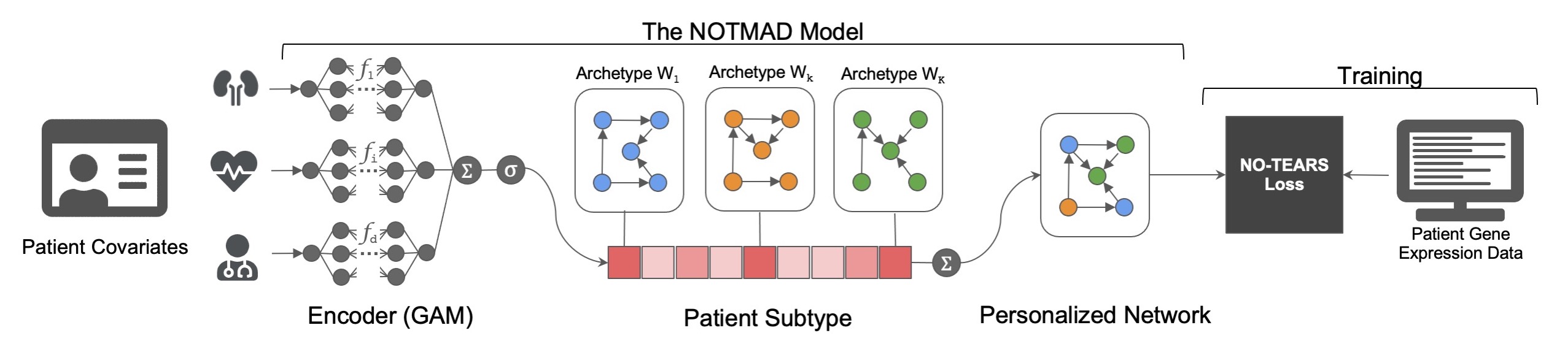}
    \caption{Our motivating example: patient-specific network inference for understanding cancer at high resolution. 
    We know that each patient's tumor is uniquely shaped by genetic and environmental factors. 
    We would like to summarize the gene regulatory relationships in each tumor as a BN to identify mutations and therapeutic targets. 
    However, the assays which measure gene expression destroy the tissue sample, so traditional methods of network inference which treat networks in isolation are unable to estimate patient-specific networks. 
    Our framework instead views each network as the output of a function of patient-specific covariates, which can be used to predict the patient's gene network as a personalized mixing of archetypal networks. 
    In this view, the generative function of regulatory networks is shared between many patients and can be learned efficiently, even when we have only a single sample from each patient.
    }
    \label{fig:framework}
\end{figure*}

Estimating the structure and parameters of Bayesian networks (BNs) is a classic problem in which we seek to summarize relationships between variables by factorizing the joint distribution as a probabilistic graphical model. 
This approach has been applied to many problems in biology \citep{sachs2005causal}, genetics \citep{zhang2013}, machine learning \citep{koller2009probabilistic}, and causal inference \citep{spirtes2000causation}. 
However, the relationships between variables are often non-stationary and context-specific. 
Previous attempts to model these context-specific relationships, such as time-varying networks \citep{song2009keller,ahmed2009recovering,parikh2011treegl}, have enjoyed success in estimating network connections which vary according to a single observed axis of variation (e.g. time); however, when dealing with complex processes which vary according to many processes simultaneously, we often do not know \emph{a priori} how observed covariates relate to network structure. 
Thus, we would like a method which can estimate BNs with structures and coefficients that vary according to several, and possibly latent, continuous factors. 
Such a method would enable us to understand large, heterogeneous datasets as directed relationships which rewire in continuously-changing contexts.

Toward this end, we propose \methodfull~\methodparens, a framework for network estimation which views BNs as the output of a learnable function of sample context. 
There are two main challenges is designing such a method: how can we enforce acyclicity (traditionally expressed as a combinatorial constraint) on the \emph{output} of a learned function, and how should we encode latent network structure in a learnable function?
Our key insight is that acyclicity function can be imposed as a smooth regularization loss on the graph generator, allowing context-specific networks to be modeled as mixtures of archetypal DAGs. 
As a result, \method~is able to generate context-specific BNs at single-sample resolution, allowing us to ask new questions regarding sample heterogeneity and latent structure. 

\paragraph{Motivating Example: Gene Regulatory Network Inference for Personalized Cancer Analysis}
A common task in bioinformatics is to infer regulatory networks from gene expression data. 
That is, we seek a graphical model which factorizes the distribution of gene expression, from which we can infer regulatory relationships. 
This graphical model representation is especially useful for cancer analysis: cancer is a highly individualized disease \citep{mourikis2018patient}, for which distinct biological processes are important for understanding and treating different tumors; moreover, cluster-based analyses are limited because tumors may not form discrete clusters \citep{ursu2020massively}. 
Thus, we seek patient-specific BNs to identify dysregulated pathways and therapeutic targets at patient-specific granularity. 
However, bulk RNA sequencing assays which measure gene expression destroy the physical sample; as a result, we are given only a single instance from which to view each tumor. 
Thus, traditional methods to estimate BNs by clustering samples are incapable of providing patient-specific inferences. 
In this paper, we develop a method (Figure~\ref{fig:framework}) which views BNs as the output of a learnable function of contextual patient information. 
This enables us to estimate personalized gene regulatory networks, and perform personalized analysis of cancer patients while sharing statistical power between many samples.

\paragraph{Contributions}
This paper contributes \method, a method for estimating context-specific BNs which enables sample-specific network inference, enabling analyses of network heterogeneity.  
Underlying \method~is a meta-view of structured models in which parameters are outputs of a shared, learnable graph generator function. 
Previously, it was thought that the structured aspect of these models prohibited graph generator functions; we show that these structured models (e.g. context-specific BNs) can be efficiently modeled as interpolations between a small number of latent archetypes, and the non-convexity of the model domain can be enforced with regularization. 
This perspective unites two areas of recent interest in machine learning: structure learning and contextualization. 
\texttt{Python} code of \method~is available at \newline \href{https://www.github.com/cnellington/SampleSpecificDAGs}{\url{https://github.com/cnellington/SampleSpecificDAGs}}. 

\section{Preliminaries}

\paragraph{DAG Structure Learning}
The basic DAG learning problem is formulated as: Let $X \in \mathbb{R}^{n \times p}$ be a data matrix consisting of $n$ IID observations of the random vector $X = (X_1,\ldots,X_p)$. Let $\mathbb{D}$ denote the space of DAGs $G = (V, E)$ of $p$ nodes. 
Given $X$, we seek weighted adjacency matrix $W \in \mathbb{R}^{p \times p}$ which defines a DAG $G \in \mathbb{D}$ with structure defined by the binarized adjacency matrix $A(W)$ where $A(W)_{ij} =1 \iff W_{ij} \neq 0$, and describes the joint distribution $\mathbb{P}(X)$ via a structural equation model \citep{spirtes2000causation,koller2009probabilistic}. 

\paragraph{Structural Equation Models}
To translate the weighted adjacency matrix $W \in \mathbb{R}^{p \times p}$ into a joint distribution on $X$, we use a structural equation model (SEM). Linear SEM models $X_j = W_j^TX + \epsilon_j$, where $X = (X_1,\ldots,X_p)$ is a random vector and $\epsilon=(\epsilon_1,\ldots,\epsilon_p)$ is a random noise vector. 
This form can be modified to use more flexible transition functions with $\mathbb{E}(X_j | X_{pa(X_j)}) = f(W_j, X)$ (e.g., for Boolean $X_j$, we can use a logistic link function). 
For the remainder of this paper, we will assume linear SEM with $\epsilon$ Gaussian and drop this specification; however, we note that \method~easily extends to other SEM by replacing this portion of loss function. 
Thus, a DAG $W$ defines a Bayesian Network (BN) distribution $\text{BN}(X| W)$ which provides the likelihood of observing $X$ under structure $W$ and imposes least-squares loss $\ell(W; X) = \frac{1}{2n}\norm{X - XW}^2_F$.

\section{Inferring Context-Specific Bayesian Networks}
\subsection{Problem Definition}

We are interested in learning BNs which are \emph{context-specific}; i.e., the parameters and/or structure of the BNs may vary according to context. 
In this problem, for each observation $X \in \R^p$, we also observe contextual data $C \in \R^m$. 
We can describe this as factorizing 
$$P(X,C) = \int_WdW P(X|W)P(W|C)P(C),$$ 
where $P(X|W) = \text{BN}(X|W)$ (recalling $\text{BN}(\cdot|W)$ was defined previously as the distribution implied by BN structure $W$). 
Drawing back to the motivating example discussed in Section 1, we can view $X$ as gene expression, $W$ as a gene regulatory network, and $C$ as patient covariates which contextualize the network. 
This is similar to a varying-coefficient (VC) description of BNs, which would model 
$W = f(C)$ with some learnable $f$. 
However, it is not tractable in practice to estimate context-specific BNs using this VC form because DAGs are high-dimensional and DAG space is non-convex, so defining $f$ to generate DAGs is unsolved. 

We are particularly interested in the extreme case where there is at most one observation per context, i.e. sample-specific BNs. 
In such cases, there is limited power to infer each BN. 
This differs from other applications in we many samples per context (e.g. $C$ is discrete and takes on finitely-many values), and cluster-based models may succeed. 
In the most challenging setting of sample-specific inference, we require a tool to share power between sample-specific estimators.

\subsection{Our Approach: Smooth Network Inference and Contextualization}
\label{sec:our_approach}

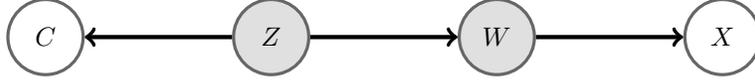
\begin{figure*}[htb]
    \centering
    \begin{tikzpicture}[
        observednode/.style={circle, draw=black!60, very thick, minimum size=10mm},
        latentnode/.style={circle, draw=black!60, fill=black!12, very thick, minimum size=10mm},
        squarednode/.style={rectangle, draw=red!60, fill=red!5, very thick, minimum size=5mm},
        ]
        \node[observednode]  at (8, 0)    (x)       {$X$};
        \node[latentnode]   at (5, 0)     (w)       {$W$};
        \node[latentnode]   at (2, 0)     (z)         {$Z$};
        \node[observednode] at (-1, 0)    (c)     {$C$};
        
        \draw[->, ultra thick] (z.west) -- (c.east);
        \draw[->, ultra thick] (z.east) -- (w.west);
        \draw[->, ultra thick] (w.east) -- (x.west);
    \end{tikzpicture}
    \caption{Graphical Model. Contextual covariates $C$ and observations $X$ are observed, while subtype $Z$ and BN parameters $W$ are latent.}
    \label{fig:pgm}
\end{figure*}

As described above, the main challenge in inference of context-specific BNs is that $W$ is a high-dimensional, structured latent variable which must satisfy acyclicity. 
To overcome this challenge, we assume that context-specific BNs lie on a subspace measured by a latent variable $Z \in \R^K$ such that $C \perp (X, W) \,\mid\, Z$. 
A graphical diagram of this meta-model is shown in Fig.~\ref{fig:pgm}. 
Drawing back to the motivating example, we can imagine that tumors correspond to latent molecular subtypes \citep{wiechmann2009presenting,le2015future} and mutations $Z$ which govern the underlying gene regulatory functions $W$; this conditional independence implies that knowing the patient covariates $C$ would not give any additional insight into the gene expression values $X$ if either the molecular subtype or the graph structure $W$ were already known.

With this meta-model, 
$\P(W|X, C)~\propto~\P(X|W)\int_ZdZ\P(W|Z)\P(Z|C)$, where $\P(X|W) = \text{BN}(X|W)$ 
and our challenge reduces to defining tractable $\P(W|Z)$ and $\P(Z|C)$. 
In \method, we use a smooth DAG criterion to define $\P(W|Z)$ and use contextual parameter generation to estimate latent $Z$ to generate $W$. 
In this way, our approach to context-specific network inference builds on two developing themes in machine learning: smooth DAG inference and contextual parameter generation.

\paragraph{Smooth DAG Inference via NOTEARS}
Estimating DAG structure, even for a single population model, from data is NP-hard \citep{chickering1996learning}, 
largely due to the acyclicity constraint which has traditionally been formulated as a combinatorial optimization. 
Recent work \citep{zheng2018dags}, however, has translated this combinatorial optimization into a continuous program through the matrix exponential: $h(W) = tr(e^{W \cdot W}) - p = 0 \iff W \in \DAG$, where $\cdot$ is the Hadamard product and $e^A$ is the matrix exponential of $A$. 
This expression of acyclicity is useful because $h(W)$ has a simple gradient: $\nabla h(W) = \left(e^{W\cdot W}\right)^T\cdot 2W$. 
In \method, we relax $h(W)$ into a regularizer which is backpropagated for gradient-based training.

\paragraph{Contextual Parameter Generation}

As datasets continue to increase in size and complexity, sample-specific inference has driven interest in many application areas
\citep{buettner2015computational,ng2015,FisherE6106,doi:10.1111/mbe.12109,ageenko2010personalized}. 
Several frameworks have been proposed to estimate meta-models which use contextual information $C$ to generate model parameters $\theta$. 
In general, we can write the operation of these frameworks as $\theta = Z^T Q$, where $Z = g_{\phi}(C)$. 
This form encompasses several well-known framework of contextualized models: if $g$ is a linear model and $Q$ is an identity matrix, we have the varying-coefficients model \citep{hastie1993varying}, if $g$ indexes the $k$-nearest neighbors in $Q$ by a learned distance metric on $C$, we have personalized regression \citep{lengerich2019learning}, if $g$ is a deep neural network and $\norm{Z} = 1$, we have Contextual Explanation Networks \citep{al2020contextual}, if $g$ is a few steps of gradient-descent beginning at the initialization $\phi$ for the instances grouped by $C$, we have meta-learning \citep{finn2017model,schaul2010metalearning}. 
In \method, $Q$ is a learned dictionary of DAG archetypes and $g$ is a differentiable context encoder (details in Sec.\ref{sec:notmad}).

\subsection{Related Work}
\label{sec:related_work}

This problem of estimating sample-specific BNs has its roots in context-specific independence (CSI) \citep{boutilier2013context}, which permitted that the independence relations captured in BN sparsity patterns may hold only for certain contexts. 
In general, CSI transforms the traditional BN DAG into a multi-graph; a long line of work have invested into developing alternative representations and approaches for estimating CSIs \citep{friedman1996discretizing,chickering1997efficient,pensar2015labeled,oates2016exact}, and more recent approaches have used causality rules to perform instance-specific independence tests \citep{jabbari2018instance,jabbari2019empirical,jabbari2020instance}. 
However, these approaches rely on the combinatorial view of acyclicity and hence suffer scalability challenges. 

There have also been approaches to design estimators of multiple BNs without sample-specific covariates. 
These frameworks include priors governing network rewiring \citep{gan2019bayesian,lee2020bayesian} and sample-specific networks as deviations from a population network \citep{kuijjer2019estimating}. 
For example, LIONESS \citep{kuijjer2019estimating} estimates linear network for sample $i$ as the weighted difference between the network estimated by all $n$ samples and the network estimated by $n-1$ samples, leaving out the $i$th sample. 
While this approach showcases the clever idea of the exchangability of these estimators and differences in linear SEM, it does not share statistical power between sample estimators (the statistical power is only shared for the population model). 
A similar approach was used to identify sample-specific dysregulated pathways in a gene regulatory network by examining the residual of neighborhood regressions \citep{buschur2019causal}, although this approach was more robust than the above because it sought only to estimate dysregulation rather than the underlying network. 
There are fundamental limits to the estimation of populations of parameters without any covariate information \citep{vinayak2019maximum}; for this reason, we concern ourselves only with the setting where we observe covariate information to contextualize the BN.

Finally, we are not the first to study mixtures of archetypal DAGs. 
Past methods have focused on estimating the latent mixture subtype of a population \citep{thiesson2013learning, strobl2019improved}, or of individual samples \citep{saeed2020causal}; however, these methods use localized structure-search algorithms, which creates a computational bottleneck that constrains learning to only mixing a pre-defined set of archetypal DAGs. 
This severely hinders the adaptability of these models because it requires a perfect prior knowledge about the causal structures underlying DAG mixtures. 
In contrast, \method~uses smooth optimization to learn archetypal DAGs without any prior knowledge.

\section{\method: \methodfull}
\label{sec:notmad}

To estimate sample-specific BNs, we propose \method, which represents context-specific distributions as: 
\begin{align}
\label{eq:bnmodel}
    \mathbb{P}(X|C) &= \text{BN}\left(X| \phi_{\theta}(C)\right),
\end{align}
which equates to $X = \phi_{\theta}(C)^TX + \epsilon$ under the linear SEM, where $\epsilon$ is a noise vector. 
Context-specific DAGs $W=\phi_{\theta}(C)$ are generated by $\phi_{\theta}$ as a convex combination of a fixed number ($K$) of archetypal DAGs (
$W_{1:K}$):
\begin{align}
\label{eq:Wmodel}
    W &= \phi_{\theta}(C) =  \sum_{k=1}^K \sigma\left(f_{\theta}(C)\right)_k \left[W_k \cdot (1 - I) \right] 
\end{align}
where $\sigma(\cdot)$ is the softmax function and $f_\theta(\cdot)$ is the context encoder, a learned function parameterized by $\theta$. 
The softmax constraint ensures that the weighting of each archetype is positive and sums to $1$, permitting these values to be interpreted as the probability of class membership. 
The dictionary of archetypal DAGs, $W_{1:K}$, is estimated jointly with the context encoder $f$, and the entire architecture is trained end-to-end. 
Finally, we clamp the diagonal of each $W_k$ to $0$ to avoid the cyclic $W=I$.

To ensure that each $W$ is acyclic, we balance the estimation loss with the NOTEARS DAG-ness loss $h(W)$. 
Thus, \method~
can be summarized as the following minimizer:
\begin{align}
    \argmin_{\theta, W_{1:K}} \sum_{i=1}^n \left(X^i - X^i\phi_{\theta}(C^i) \right)^2 + \alpha \text{tr}\left(e^{\phi_{\theta}(C^i) \cdot \phi_{\theta}(C^i)}\right) + \sum_{k=1}^K\beta \norm{W_k}_1 + \gamma\text{tr}\left(e^{W_k \cdot W_k}\right)
\end{align}
where $\alpha$, $\beta$ and $\gamma$ are hyperparameters that trade off predictive loss against DAG-ness and sparsity, and we recall that $h(W) = \text{tr}(e^{W \cdot W})$ is the acyclicity regularizer discussed in Section~\ref{sec:our_approach}, here applied to both the context-specific and the archetypal DAGs. 
As introduced in Sec.~\ref{sec:our_approach}, \method~has a natural probabilistic interpretation. 
We have defined $\E[\P(W|Z)] = \sum_k Z_k[W_k \cdot (1 - I)]$, and constrained $Z \in \{q \in \R_{\geq 0}^{K} : \norm{q} = 1\}$ by $Z=\sigma(f_{\theta}(C))$ such that these weightings are a convex combination.
\texttt{Python} code of \method~is available at \href{https://www.github.com/cnellington/SampleSpecificDAGs}{\url{https://github.com/cnellington/SampleSpecificDAGs}}. 

\paragraph{Context Encoder}
This definition of \method~is flexible with respect to the form of the context encoder $f$ -- any differentiable function can be used. 
We have found best success and report all experimental results using a Neural Additive Model \citep{agarwal2020neural}, which rivals the accuracy of fully-connected neural networks while retaining interpretability. 
For other types of contextual data, other encoder models may be more appropriate; e.g. a convolutional neural network for images.

\paragraph{Fine-Tuning}
The NOTEARS regularizer encourages the context encoder to produce acyclic graphs; however, we are not guaranteed that the graph predicted by \method~will be acyclic for all unseen test contexts. 
Nevertheless, any directed graph can be projected to a DAG by removing edges which violate acyclicity: for fixed $W$, we perform a binary search over the magnitudes in $W$ to find the minimum value $t$ for which $W\cdot (\abs{W}>t) = W' \in \DAG$. 
This produces a DAG $W'$, to which we attempt to add back excluded edges from $W$ in order of decreasing magnitude of edge weights, checking that each edge addition does not form a cycle.
In practice, predictions from \method~do not require fine-tuning beyond thresholding out weak edges.

\paragraph{Benefits of Modeling Networks as Mixtures of Archetypes}
In some sense, \method~is comparable to a multi-output varying-coefficients model with an extra loss to encourage acyclicity. 
However, \method~uses archetype mixing to create a bottleneck and reduce the effective dimensionality of the outputs (borrowing the dictionary-learning framework of \cite{al2020contextual}). 
This bottleneck has statistical benefits: constraining the sample-specific DAGs to lie in a $K$-dimensional polytope defined with the $K$ archetypes as extreme points. 
In addition, this has computational benefits: it is not tractable to store many $p \times p$ matrices in memory; the archetype view allows efficient computation and optimization. 
Finally, archetypes aid interpretability: we can understand latent membership patterns by visualizing the context-subtype mapping in the additive context encoder $f_{\theta}$.

\paragraph{Non-Convexity}
Is it reasonable to define sample-specific DAGs as convex combinations of archetypal DAGs? 
The DAG constraint is non-convex, so it is not true that the convex combination of any set of DAG archetypes would be acyclic.  
If DAG space is highly disconnected, then we should not expect optimization to efficiently learn to generate DAGs by combining archetypal DAGs. 
Is \method~hopeless?

Inspecting the topology of DAG space gives hope: the compatibility of DAGs is driven by the sparsity structure of summands, not the weighting.
Let $A(W)$ be the binarized structure of $W$ (i.e. $A(W)_{ij} =1 \iff W_{ij} \neq 0$). Then $A(W_1)+A(W_2) \in \DAG \implies aW_1 + W_2 \in \DAG \forall ~a$. 
Conversely, if $A(W_1)+A(W_2) \notin \DAG$, then $aW_1+W_2 \in \DAG$ for at most $p^2$ values of $a$. 
As a result, there is minimal confusion over allowable archetype combinations: either the combination is acyclic for all weightings, or there is a small, finite set of weightings which produce DAGs and all other weightings produce cyclic graphs. 
Thus, the context encoder $f$ needs only to learn compatible archetypes, not compatible weightings.

\section{Experiments}
\label{sec:experiments}
BNs factorize joint distributions into tractable conditional dependencies; thus inference of context-specific BNs provides insight into the changing dependencies in heterogeneous data. 
Traditionally, structure learning has required multiple observations for each network; in contrast, \method~enables us to estimate sample-specific networks for unclustered individual observations. 
Here, we evaluate the success and limitations of \method~to identify the accuracy and utility of sample-specific network inference. 

\paragraph{Baselines}
We compare the performance of \method~against estimators of BNs at various resolutions: state-of-the-art population estimator (NOTEARS \citep{zheng2018dags}), a cluster-specific estimator which first clusters the samples according to the context and then applies the population estimator to each cluster, and sample-specific BN inference by LIONESS \citep{kuijjer2019estimating}. 
Finally, when we have domain knowledge regarding latent sample structure, we construct ``oracle''-clustered estimators by clustering the samples according to domain knowledge and applying the population estimator to each cluster.
On real data, we do not have access to the underlying BN; thus, we evaluate the mean-squared error (MSE) of $\norm{X-XW}_2$ on held-out samples which corresponds to the likelihood of the observed data under the probabilistic interpretation of the linear SEM.

\begin{table*}[b]
    \begin{center}
    \begin{tabular}{c|c|c|c|c|c}
    \bf{Dataset} & \bf{Population} & \bf{Clustered} & \bf{Oracle Clustered} & \bf{LIONESS} & \bf{\method} \\ \toprule
        \bf{Finance} & $0.58 \pm 0.01 $ & $0.54 \pm 0.01 $ & -*- & $0.67 \pm 1.94$ & $\bf{0.43 \pm 0.008}$ \\
        \bf{Football} & $0.47 \pm 0.006$ & $0.44 \pm 0.006$ & $0.41 \pm 0.004$ & $0.52 \pm 0.19$ & $\bf{0.35 \pm 0.005}$\\
        \bf{Cancer} & $0.49 \pm 2.3\mathrm{e}{-5}$ & $0.37 \pm 6.5\mathrm{e}{-4}$ & $0.24 \pm 1.1\mathrm{e}{-4}$ & $0.65 \pm 0.14$ & $\mathbf{0.14 \pm 2.7\boldsymbol{\mathrm{e}{-5}}}$ \\
        \bf{Single-Cell} & $0.56 \pm 0.02$ & $0.45 \pm 0.01$ & $0.37 \pm 0.02$ & $0.60 \pm 0.08$ & $\bf{0.28 \pm 0.02}$ \\
    \end{tabular}
    \caption{Mean-squared error (MSE) loss of BNs applied to held-out samples. Values displayed are the mean $\pm$ variance of the MSE over 10 bootstrap samples. *No oracle for clustering.
    \label{tab:mse_results}}
    \end{center}
\end{table*}

\subsection{Finance} 
Finally, we explore heterogeneity in a dataset of financial asset prices and news headlines from \cite{lengerich2019learning}. 
This dataset was proposed to show the utility of allowing regression parameters to change over time and industry; here we examine how the inter-relatedness of the samples drifts over time by estimating contextualized BNs. 

For each date, we have as context $C$ a summary of the news headlines as a 50-dimensional vector, while the predictors $X$ are the trading price of 24 assets. 
We split the 14,555 samples into train/test sets at the 75th percentile date, which is approximately the beginning of 2011, to produce a training set of 10,916 samples. 
From these data, we estimate daily BNs of asset price relations contextualized by history and news. 

\begin{figure*}[h]
    \centering
    \begin{subfigure}[b]{0.47\textwidth}
        \centering
        \includegraphics[width=0.45\textwidth]{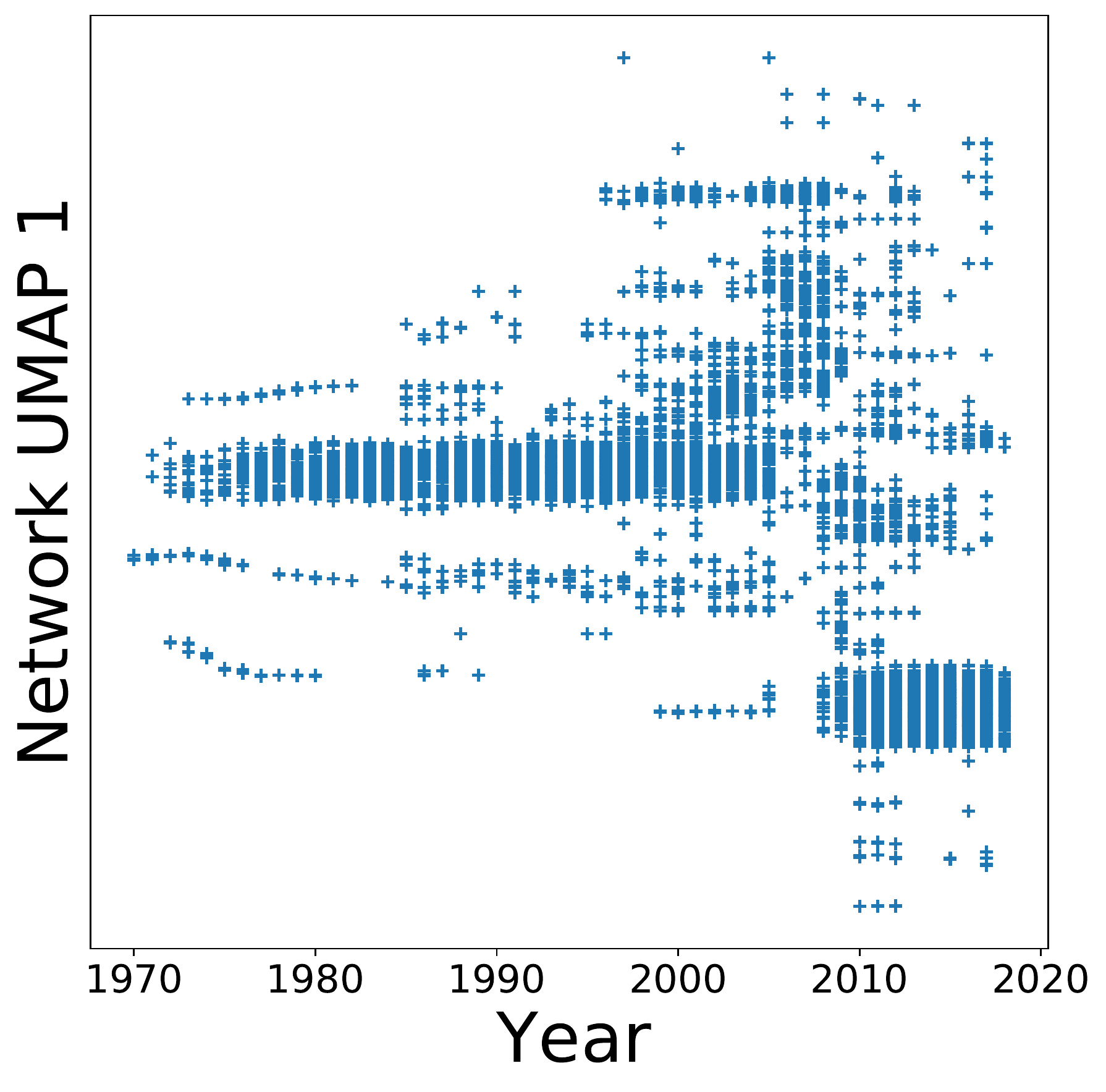}
        \caption{Embeddings of context-specific BNs by date. Context-specific BNs recover trends leading to and the broad dislocation after the 2008 financial crisis.}
        \label{fig:finance_by_year}
    \end{subfigure}
    ~
    \begin{subfigure}[b]{0.5\textwidth}
        \centering
        \includegraphics[width=0.55\textwidth]{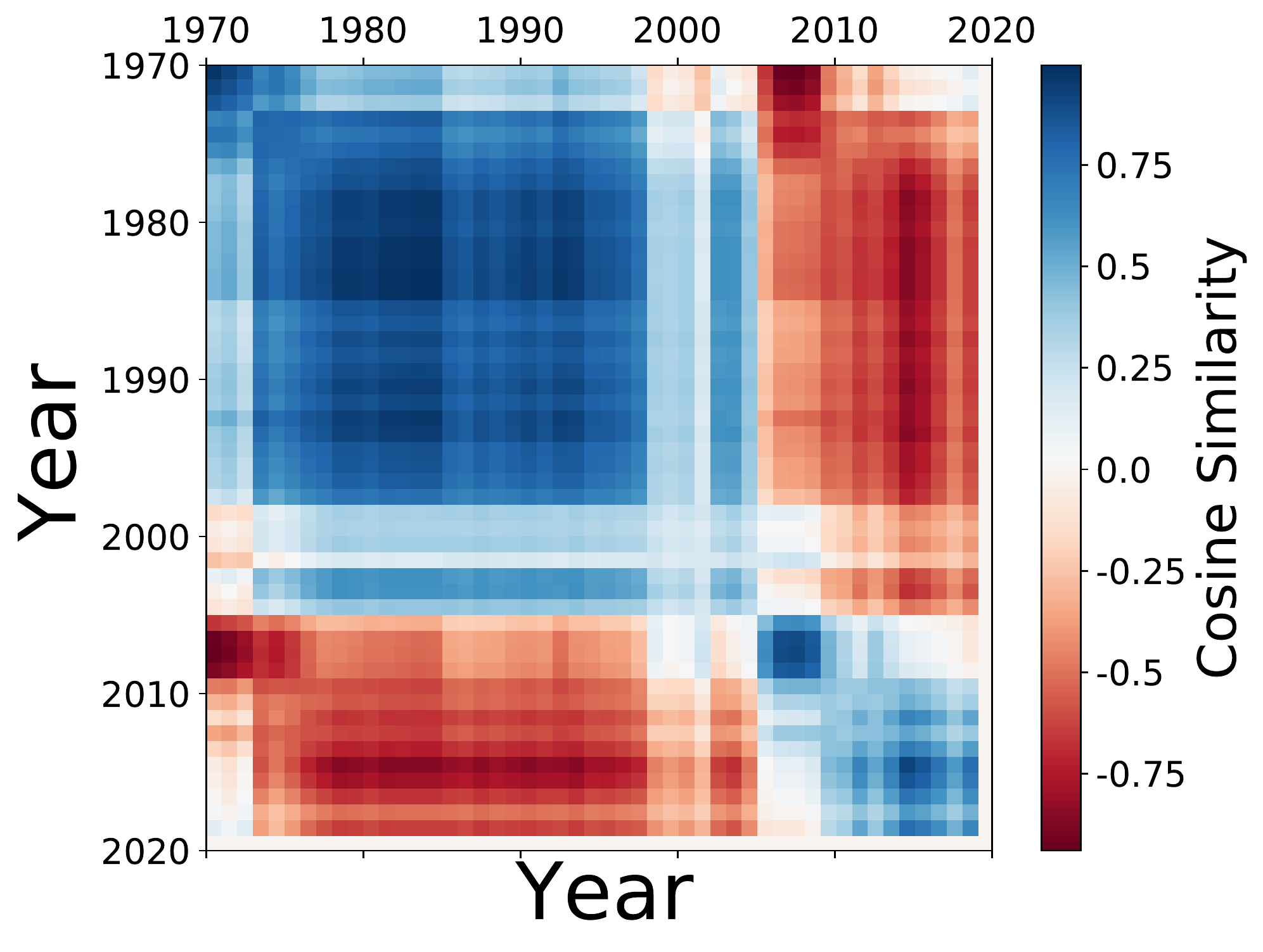}
        \caption{Cosine similarities between context-specific BNs by date. Context-specific BNs recover macro trends including 1970s stagflation, 2000s growth, and 2008 financial crisis.}
        \label{fig:finance_similarities}
    \end{subfigure}
    \caption{Representations of context-specific BNs estimated for financial asset prices.}
    \label{fig:results_finance}
\end{figure*}

In addition to giving the best MSE loss on held-out samples (Table~\ref{tab:mse_results}), the estimated BNs show interpretable patterns. 
As shown in Fig.~\ref{fig:results_finance}, the sample-specific networks accurately translate the changing environment into interpretable networks of financial relationships.

\subsection{Football}
A goal in sports analytics is to isolate the influence of individual players from the abilities of teammates. 
Professional players are often rated by independent rating agencies; we can these use ratings to identify influences by estimating a BN to factorize player ratings. 
In such a BN, each node represents a player rating while edges represent how the rating of a player is influenced by the ratings of teammates. 
Context is critical: team strategies dictate player expectations and analyses. 
We use ratings\footnote{\url{https://www.lineups.com/nfl/player-ratings}} of 1000 professional football players from the 2020 National Football League season. 
We group players into 7 position categories, and construct simulated observations of lineups by team.  
For context, we use the team identity and a PCA-reduced representation of the ratings; for oracle clustering, we use team identity. 
In this dataset, $n=1478$ (with 634 test samples), $p=7$, and $m=4$. 
As shown in Table~\ref{tab:mse_results}, context-specific BNs capture the distributions of player ratings more accurately than population or cluster-based BNs.

\subsection{Gene Regulatory Network Inference for Personalized Analysis of Cancer}
\label{sec:experiments:gene}
We now return to the motivating example discussed in Section~\ref{sec:introduction}.

\paragraph{Dataset}
We use bulk RNA-seq gene expression data from The Cancer Genome Atlas \footnote{\url{https://portal.gdc.cancer.gov/}} (TCGA), which measures the average expression of each gene in a patient's tumor. 
We select 784 RNA-seq transcripts (features) based on annotation in the COSMIC gene census \cite{forbes2014cosmic} and large variance over the sample.
These gene expressions levels are measured for 9715 patients, which each have 12 contextual covariates: Age, Gender, Race, Sample Type, Primary Site, \% stromal cells, \% tumor Cells, \% Normal Cells, \% Neutrophil Infiltration, \% Lymphocyte Infiltration, \% Monocyte infiltration, and a single principal component of gene expression which captures batch effects in processing. 
The cancers span 20 tissues and 26 types of tumors.

\begin{figure}[htbp]
    \centering
    \includegraphics[width=\textwidth]{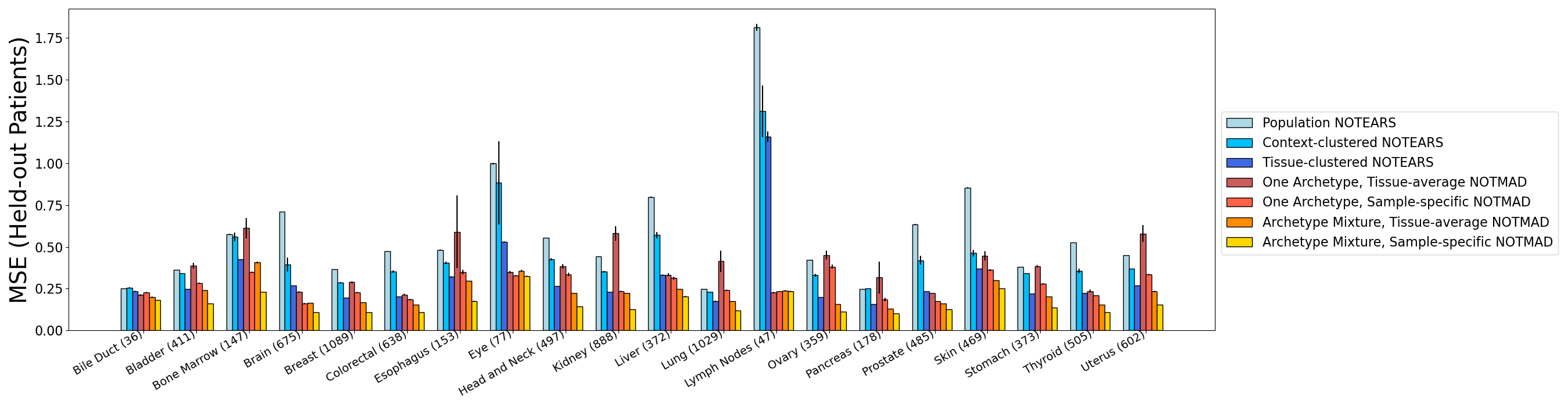}
    \caption{MSE of BNs of gene expression in cancer, grouped by tissue. The number of training samples for each tissue type is provided in parentheses. In addition to the population, cluster-specific, and tissue-specific baselines, we also include ablations of the \method~model: ``One Archetype, Tissue average'' uses the power sharing of \method~across tissues, but only estimates a single BN for each tissue; ``Archetype Mixture, Tissue average'' similarly estimates a single BN for each tissue but allows this tissue-specific BN to be a mixture of archetypal DAGs; ``One Archetype, Sample-Specific'' estimates patient-specific BNs that are constrained to use only a single archetype; ``Archetype Mixture, Sample-Specific'' is the entire \method~model.
    \label{fig:tcga_mse}}
\end{figure}

\paragraph{Results}
Patient-specific BNs estimated by \method~represent gene expression better than do population, cluster-specific, or tissue-specific BNs. 
Not only do patient-specific BNs improve average MSE, (Table~\ref{tab:mse_results}) the patient-specific BNs also improve estimation in every tissue (Fig.~\ref{fig:tcga_mse}). 
The largest out-performance is seen for cancers of the lymph nodes (MSE of tissue-specific BNs is 4.9x that of \method), brain (2.4x), thyroid (2.0x), prostate (1.9x), colon (1.8x), and head/neck (1.8x).

The high resolution of these patient-specific BNs unlocks new investigations of network heterogeneity. 
For example, are the patient-specific distributions of gene expression best viewed as a patient-specific \emph{selection} of a distribution, or a patient-specific \emph{mixture} of archetypal distributions (i.e. do gene regulatory networks in cancer rewire according to a small number of fixed processes, or do they interpolate between processes)? 
With \method, we can begin to answer this question. Here, we examine this question through an ablation study: constraining each patient-specific BNs to use only a single archetype (``One Archetype'' in Fig.~\ref{fig:tcga_mse}) or to use only a single BN for all tumors of a particular tissue (``Tissue-average'' in Fig.~\ref{fig:tcga_mse}). 
According to MSE, the full patient-specific heterogeneity of \method~is required to achieve optimal summarization of gene expression distributions, suggesting that these cancers exhibit a diversity of expression patterns that cannot be resolved by a ``cancer'' BN (i.e. a population model), tissue-specific BNs, or patient-specific BNs which are selections of clustered networks; instead, tumors as best modeled as interpolations between latent processes.

\begin{figure*}[htb]
    \centering
    \includegraphics[width=0.9\textwidth]{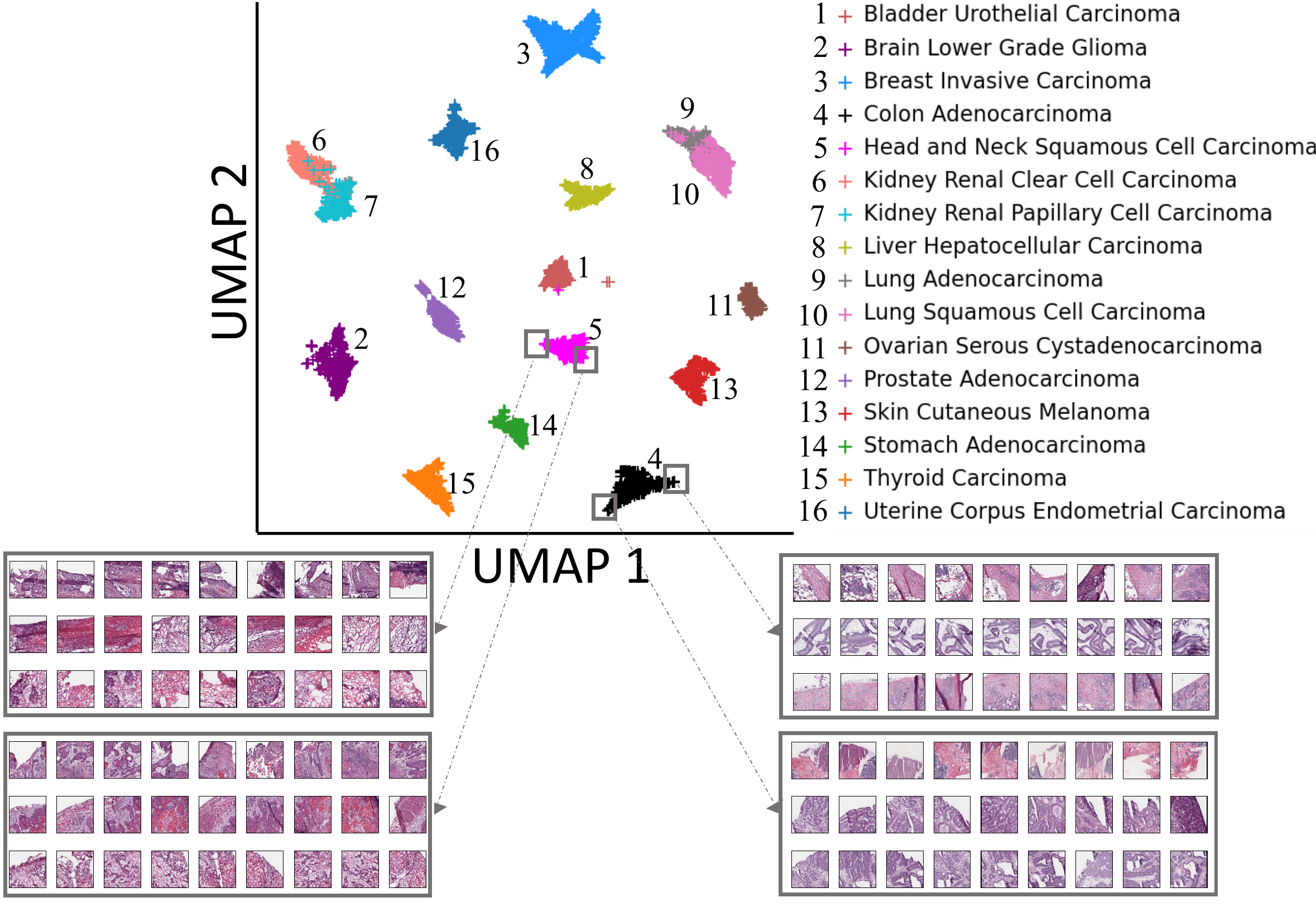}
    \caption{Patient-specific BNs embedded by UMAP. (Top) Without access to disease type during training, the patient-specific BNs accurately recover heterogeneity between cancers of disparate tissues, and cancers of different subtypes affecting the same tissue. (Bottom) Network archetype selection suggests network heterogeneity in both head and neck squamous cell carcinomas (left) and colon adenocarcinomas (right). For each archetype, we show image patches from the 3 cases (by row) which maximize \method~activation of the archetype (10 patches each) and draw arrows to the corresponding sample group containing these cases. 
    \label{fig:tcga_embeddings}}
\end{figure*}

In addition, the patient-specific BNs recover medically-significant heterogeneity. 
By inspecting network embeddings (Fig.~\ref{fig:tcga_embeddings}, we see that the BNs estimated by \method~tend to cluster according to tissue. 
Nevertheless, the patient-specific BNs separate lung adenocarcinoma (LUAD) samples from lung squamous cell carcinomas (LUSC). 
These subtypes of lung cancer have important distinctions in clinicopathology (e.g LUSC is associated with a history of smoking, while LUAD is the most common form of lung cancer in non-smokers \citep{Relli2019}) and treatment (e.g. epidermal growth factor receptor and anaplastic lymphoma kinase are two commonly mutated genes in LUAD and are targets for current therapies, while they are infrequently mutated in LUSC \citep{Lindeman2013}). 
Since \method~did \emph{not} have access to this LUAD/LUSC distinction as contextual data, this distinction must be a result of optimizing MSE loss, indicating different regulatory networks underlying these biomedically-distinct subtypes of lung cancers. 

In addition to recovering known heterogeneity of disease types, \method~also identifies intra-disease heterogeneity which correspond to morphological distinctions. 
For example, the BNs for colon cancers tend to be interpolations between two archetypal networks, and similarly head and neck cancers have two archetypal networks underlying the patient-specific BNs. 
For colon cancer, this heterogeneity may correspond to the four consensus molecular subtypes \cite{guinney2015consensus}, while for head and neck cancers, this split may correspond to differences in head and neck tumors arising from either tobacco use and/or HPV \citep{Stransky2011-nw}.  
Whole-slide imaging of these tumors (Fig.~\ref{fig:tcga_embeddings}) shows the morphological distinctions between the two latent BN archetypes, suggesting a correspondence between gene regulatory network heterogeneity and tumor morphology.

\begin{figure}[htp]
    \centering
    \begin{subfigure}[b]{\textwidth}
        \centering
        \includegraphics[width=0.85\textwidth]{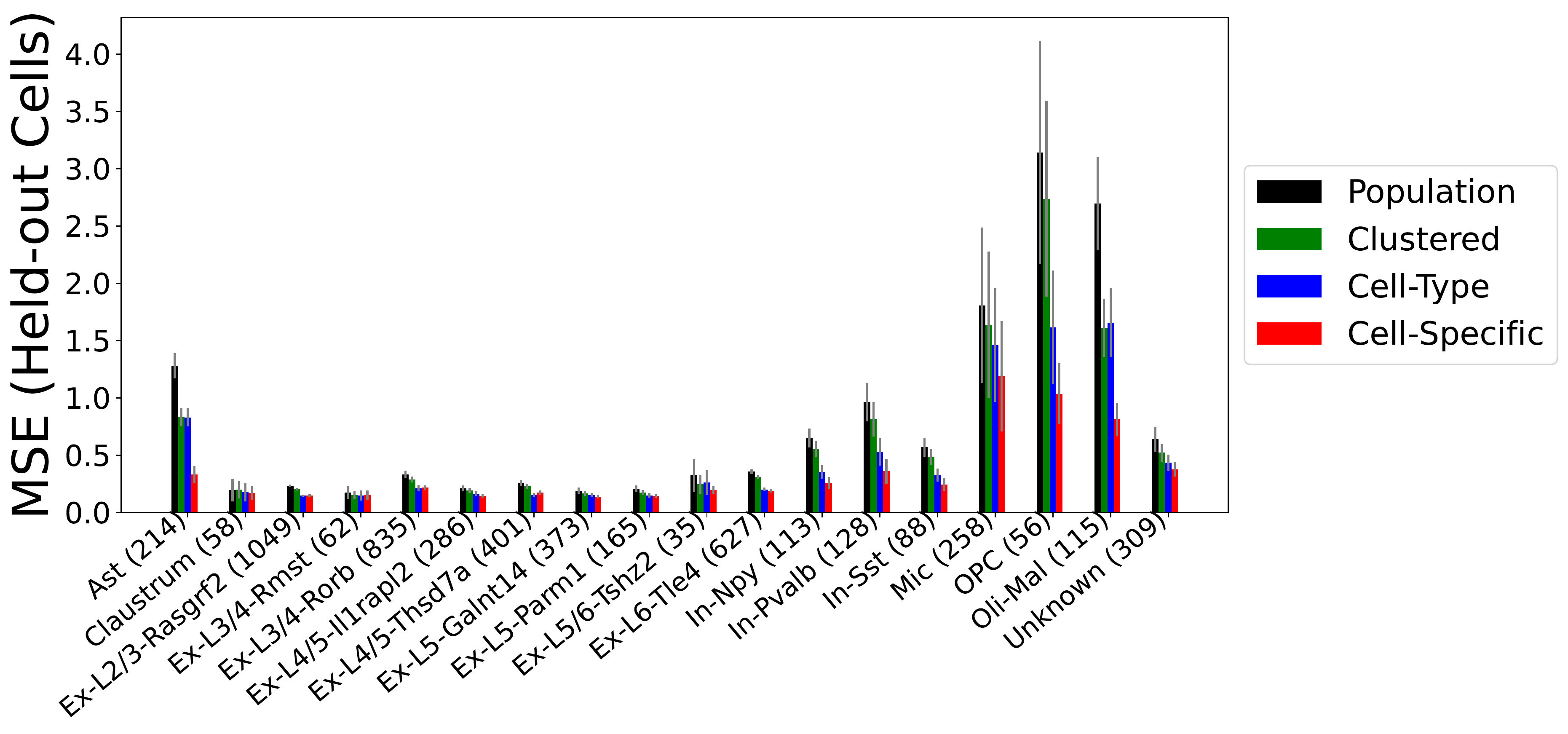}
        \caption{Mean-squared error (MSE) of BNs of single-cell gene expression. We compare four models: (1) a population model which estimates a single BN, (2) a clustered model which estimates a small number of discrete clusters of samples and cluster-specific BNs, (3) cell-type specific networks which a BN for each cell type, and (4) cell-specific networks estimated with NOTMAD. In all cell types, the cell-specific networks estimated by NOTMAD are the most accurate.}
    \end{subfigure}
    ~
    \begin{subfigure}[b]{0.48\textwidth}
        \centering
        \includegraphics[width=0.95\textwidth]{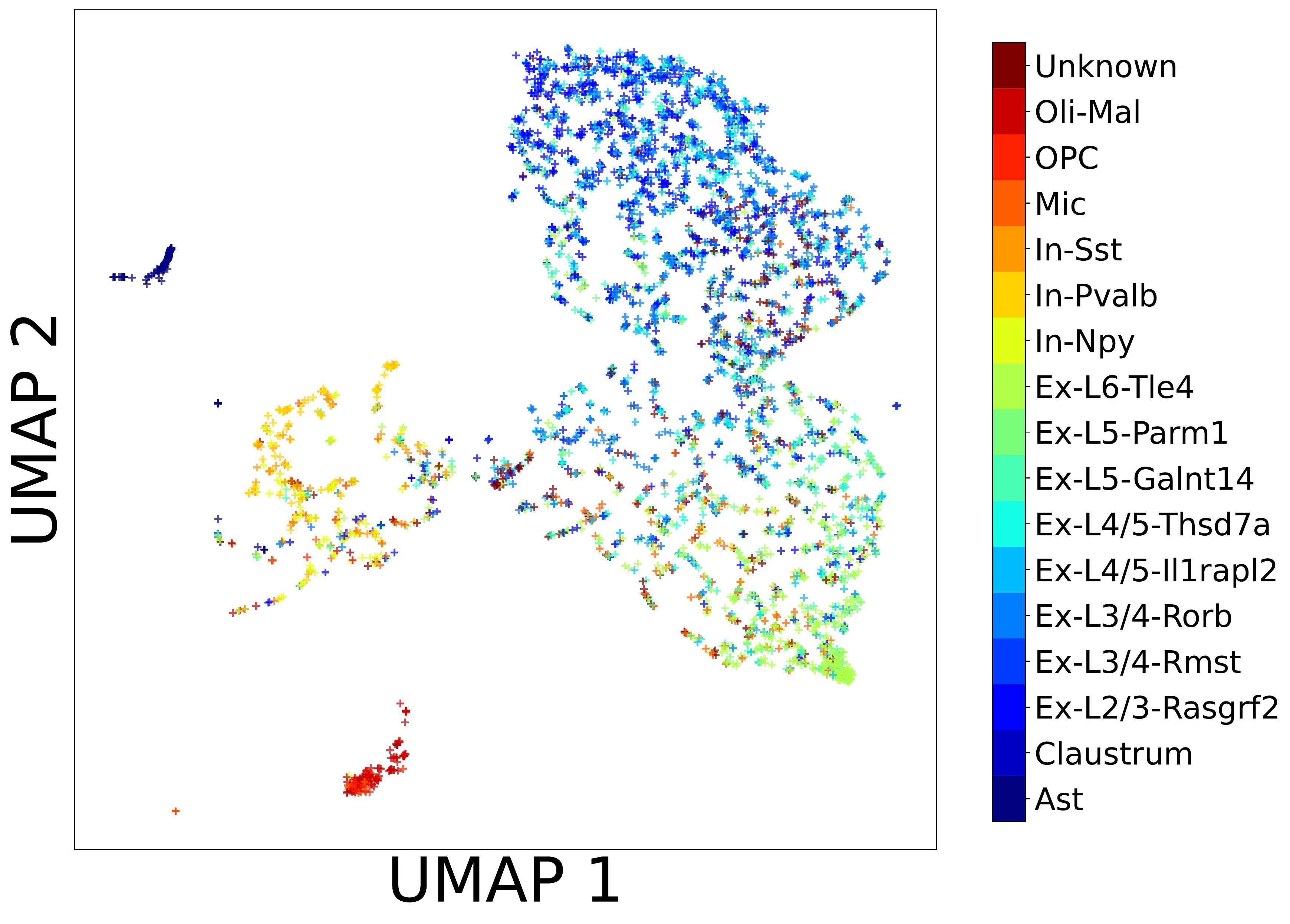}
        \caption{Network embeddings, colored by cell type. Each point represents the BN for a single cell, embedded with UMAP. The networks for inhibitory and excitatory neurons tend to cluster according to their corresponding type, with subtypes providing a modest sub-clustering. Astrocytes form a distinct cluster of networks, while Oli-Mal, OPC, and Mic cells form a separate cluster.}
    \end{subfigure}
    ~
    \begin{subfigure}[b]{0.5\textwidth}
        \centering
        \includegraphics[width=0.66\textwidth]{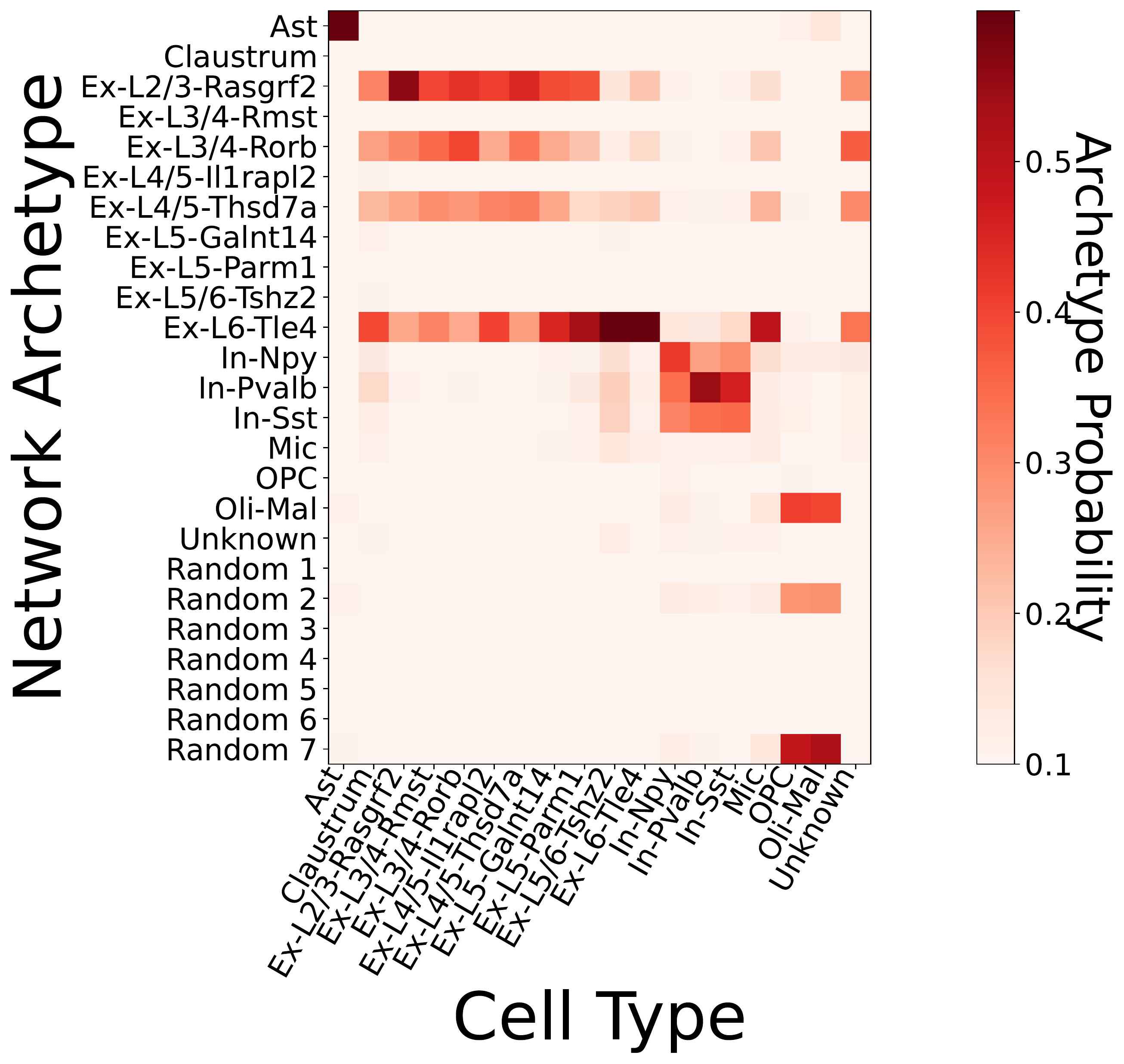}
        \caption{Network archetype selection by cell type. Color corresponds to the mean weight of each archetype (interpreted as a probability) in constructing cell-specific networks for cells of each cell type. In addition to cell type-specific network archetypes, we also initialize 7 archetypes to uniformly random sparse networks (``Random''). Excitatory and inhibitory neurons share network archetypes between subclasses, but other cell types use distinct archetypes.}
    \end{subfigure}
    \caption{Cell-specific networks reveal cell-specific heterogeneity in adult mouse brain.}
    \label{fig:snare_seq_results}
\end{figure}

\subsection{Single-Cell Gene Regulatory Network Inference}

The gene regulatory network of each brain cell is unique, shaped by its cell type, cell state, history of activation, interconnections, environment, and by the age, sex, and disease status of each person. 
However, these networks are highly related across cells and highly structured, as they are shaped by circuitry constraints, a limited set of regulatory programs, and a finite set of active regulators. 
Here, we exploit this structure by using NOTMAD to simultaneously estimate network archetypes and the cell-specific networks using paired epigenetic and transcriptomic data. 
Importantly, we learn both membership vectors and archetypal networks \emph{de novo} in an end-to-end joint optimization, in essence inferring a functional transformation that predicts a cell’s gene regulatory network from its DNA accessibility profile. 

\paragraph{Dataset}
To estimate single-cell regulatory networks, we use recently-developed dual assay technology, which allows simultaneous profiling of transcriptomic (RNA-seq) and epigenetic (ATAC-seq) markers in matched single cells. 
In this example, we use the SNARE-seq dataset of 10,309 cells from adult mouse brain cells \cite{chen2019high}. 
To preprocess the scRNA-seq, we initial perform quality control with poor quality cell removal and doublet removal. Next, we remove empty droplets using \texttt{DropletUtils} and doublets using \texttt{scds} with a threshold of $1.0$. Finally, we normalize reads with \texttt{linnorm}. Finally, we perform a PCA transform of both the scATAC-seq and scRNA-seq data to compress each to 5 covariates. 
The pre-processed dataset is available at \href{https://www.github.com/blengerich/SnareSeq}{\url{https://github.com/blengerich/SnareSeq}}.

\paragraph{Results}
As shown in Figure~\ref{fig:snare_seq_results}, cell-specific BNs more accurately capture the heterogeneity of gene expression than do cell type-specific BNs, cluster-specific BNs, or a single population BN. 
This benefit is observed for all cell types, with most benefit for cell types which are less common in the training set or more uniquely differentiated due to their biological function, suggesting that NOTMAD improves performance through both sharing power between samples and capturing heterogeneity where appropriate.  
In addition, the cell-specific networks cluster according to the cell's biological function; networks for all subtypes of inhibitory neurons cluster together, while networks for all subtypes of excitatory neurons also loosely cluster together. 
In contrast, networks for OPC, Oli-Mal, Mic, and Ast cells do not cluster with either the excitatory or inhibitory neurons.

\section{Conclusions}
We have presented \method, a method to estimate context-specific BNs by learning to transform contextual data into BN parameters. 
Previously, the structured constraints of BNs prohibited graph generator functions; we show that these structured models can be efficiently modeled as interpolations between a small number of latent archetypes, and the non-convexity of the model domain can be enforced with regularization.
As a result, \method~enables simultaneous inference of sample-specific and archetypal BNs without any prior knowledge of BN structure, parameters, or sample relationships. 
\method~scales to the extreme case of estimating sample-specific BNs, in which each context is observed for only a single statistical sample. 
BNs are a mainstay of analysis in many fields including biology, medicine, social sciences, and beyond; context-specific BNs enable new questions of heterogeneity of latent structure and provide axes of variation which can be used to describe community composition and influences at a new granularity.

\subsubsection*{Acknowledgements}
B.L. was supported by the Alana Fellowship at MIT.
C.E. and E.X were supported by NIH NIGMS R01GM140467. 
B.A. was supported by NSF IIS-1956330, NIH R01GM140467, and the Robert H. Topel
Faculty Research Fund at the University of Chicago Booth School of Business. 


\end{document}